\documentclass{article}

\PassOptionsToPackage{numbers,sort&compress}{natbib}
\usepackage[preprint]{neurips_2026}

\usepackage[utf8]{inputenc} %
\usepackage[T1]{fontenc}    %
\usepackage{hyperref}       %
\usepackage{url}            %
\usepackage{booktabs}       %
\usepackage{amsfonts}       %
\usepackage{nicefrac}       %
\usepackage{microtype}      %
\usepackage{xcolor}         %

\usepackage{tikz}
\usetikzlibrary{positioning,arrows.meta}

\usepackage{amsmath}
\usepackage{amssymb}
\usepackage{mathtools}
\usepackage{amsthm}

\usepackage{algorithm}
\usepackage{algorithmic}

\usepackage[capitalize,noabbrev]{cleveref}

\theoremstyle{plain}
\newtheorem{theorem}{Theorem}[section]

\theoremstyle{definition}

\theoremstyle{remark}
\newtheorem{remark}[theorem]{Remark}

\usepackage{mathtools}

\usepackage{pifont}

\usepackage{xcolor}
\usepackage{listings}
\usepackage[most]{tcolorbox}
\usepackage{needspace}
\usepackage{placeins}
\newtcolorbox{promptbox}[1][]{
  enhanced,
  width=\linewidth,
  colback=blue!3!white,
  colframe=blue!20!gray,
  boxrule=0.4pt,
  arc=2pt,
  left=8pt,
  right=8pt,
  top=4pt,
  bottom=4pt,
  fontupper=\small,
  before skip=0.5\baselineskip,
  after skip=0.5\baselineskip,
  #1
}
\definecolor{codebg}{rgb}{0.97,0.98,1.00}
\definecolor{codekey}{rgb}{0.10,0.30,0.70}
\definecolor{codestr}{rgb}{0.55,0.10,0.55}
\definecolor{codecomment}{rgb}{0.20,0.55,0.30}
\definecolor{codeframe}{rgb}{0.35,0.45,0.65}
\lstdefinestyle{agentsmd}{
  backgroundcolor=\color{codebg},
  basicstyle=\ttfamily\footnotesize,
  keywordstyle=\color{codekey}\bfseries,
  commentstyle=\color{codecomment}\itshape,
  stringstyle=\color{codestr},
  morekeywords={tool,name,description,parameters,type,object,properties,required,returns,plug-and-play,skill,docker,run,fetch,solve,verify,generate},
  morecomment=[l]{\#},
  morestring=[b]",
  frame=single,
  rulecolor=\color{codeframe},
  framesep=4pt,
  xleftmargin=4pt,
  xrightmargin=4pt,
  showstringspaces=false,
  breaklines=true,
  columns=fullflexible,
}
\lstdefinestyle{promptblock}{
  backgroundcolor=\color{codebg},
  basicstyle=\ttfamily\footnotesize,
  keywordstyle=\color{codekey}\bfseries,
  commentstyle=\color{codecomment}\itshape,
  morekeywords={SYSTEM,USER,ASSISTANT,VERDICT,AGREE,DISAGREE,REASON,FINAL_ANSWER},
  morecomment=[l]{\#},
  frame=single,
  rulecolor=\color{codeframe},
  framesep=4pt,
  xleftmargin=4pt,
  xrightmargin=4pt,
  showstringspaces=false,
  breaklines=true,
  columns=fullflexible,
}

\graphicspath{{./figs/}{../}}

\title{A$^{2}$utoLPBench: An Auto-Generated, Agent-Friendly LP Benchmark via Inverse-KKT Construction}

\author{%
    Shuo Ren \quad Yaohui Han \quad Yifan Shi \quad Libo Shen \\[0.4em]
    \textbf{Haodong Lu \quad Dongfang Wu \quad Rongliang Fu \quad Bei Yu \quad Tsung-Yi Ho} \\[0.4em]
    The Chinese University of Hong Kong \\
}

\begin{document}

\maketitle

\begin{abstract}
Most LP-from-text benchmarks are static datasets of word problems written and labeled by hand. Once such a dataset is released, its size is fixed, its difficulty is fixed, and every problem can leak into the training data of future LLMs.
We present \textbf{A$^{2}$utoLPBench}\renewcommand{\thefootnote}{\fnsymbol{footnote}}\footnote{Open-sourced at \url{https://anonymous.4open.science/r/AutoLPBench/}; usage and agent plug-in instructions are in \Cref{app:docker}.}\setcounter{footnote}{0}\renewcommand{\thefootnote}{\arabic{footnote}}, a benchmark for testing LLM-driven agents on linear programming problems written in plain text. It has two parts, Auto and Agent: a generator that produces fresh problems on demand, and a ready-to-run evaluation environment that any LLM-driven agent can plug into.
The generator builds each LP problem in reverse. We first pick a feasible point and dual, then write down a problem for which that point is optimal and the objective value is known. The answer is known by construction, with no solver call and no human annotator. The evaluation environment bundles a reference solver-critic baseline and a Docker image whose usage instructions are written for an
LLM-driven agent to read. With these in place, any agent can run the benchmark and get a calibrated score with one command.
Because the benchmark is a generator rather than a fixed dataset, it has properties no fixed dataset can match: an unlimited supply of fresh problems, a difficulty knob set by $(n,m)$, ground-truth answers correct by construction, low LLM-side cost per problem relative to human authoring, repeatable scores across independent batches, and resistance to training-data leakage when fresh post-cutoff seed ranges are used.
We show empirically that solve rates change smoothly with problem size,
that independent batches give matching scores, and that the bundled
solver-critic baseline also helps on existing LP benchmarks.
\end{abstract}

\section{Introduction}
\label{sec:intro}

Large language models (LLMs) have advanced rapidly toward agent-style code execution and tool use~\citet{deepseek_v32_2025,deepseek_v4_2026,qwen35_2026,kimi_k25_2026,Openai_chatgpt,Anthropic2024agents}.
This shift has created a need to measure not the surface quality of LLM output but the end-to-end problem-solving capability of an LLM-driven agent on concrete tasks. A growing body of agent-style benchmarks targets this need across different task families~\cite{liu2023agentbench,qin2023toolllm,jimenez2023swe}.

Linear programming (LP) is one such task family. An LP-from-text agent~\cite{ramamonjison2022nl4opt,huang2024mamo,yang2024optibench,ahmaditeshnizi2023optimus} receives a problem description in natural language, builds the mathematical formulation, generates solver code, executes it, and reports the optimum. Formulation, code generation, and numerical fidelity must all be correct simultaneously, which makes LP-from-text a strict end-to-end test of LLM-agent capability. 
Reliable evaluation of these agents demands a test set of problem--answer pairs whose answers are verified correct. Producing such pairs is laborious: benchmarks such as NL4OPT~\cite{ramamonjison2022nl4opt} and MAMO~\cite{huang2024mamo} each assembled hundreds of human-authored word problems whose gold answers were labeled by hand, a process that is slow, costly, and difficult to scale.

Beyond their production cost, these benchmarks share a deeper structural disadvantage: they are all released as \textbf{static corpora}, a fixed set of human-authored word problems with gold answers attached.
This design has four structural limitations that no amount of careful curation can remove:
(i) Each item requires human authoring and validation, so the benchmark size is bounded by the annotator budget.
(ii) Every released item enters subsequent training corpora; once a model is trained on the public release, the benchmark becomes a memorization probe rather than a capability test.
(iii) The difficulty distribution is fixed at release time and cannot be re-targeted as model capability changes.
(iv) Evaluation is run by hand or with paper-specific harnesses; there is no agent-runnable surface that an LLM-driven agent can plug into out of the box.
\begin{figure}[!t]
    \centering
    \includegraphics[width=0.78\linewidth]{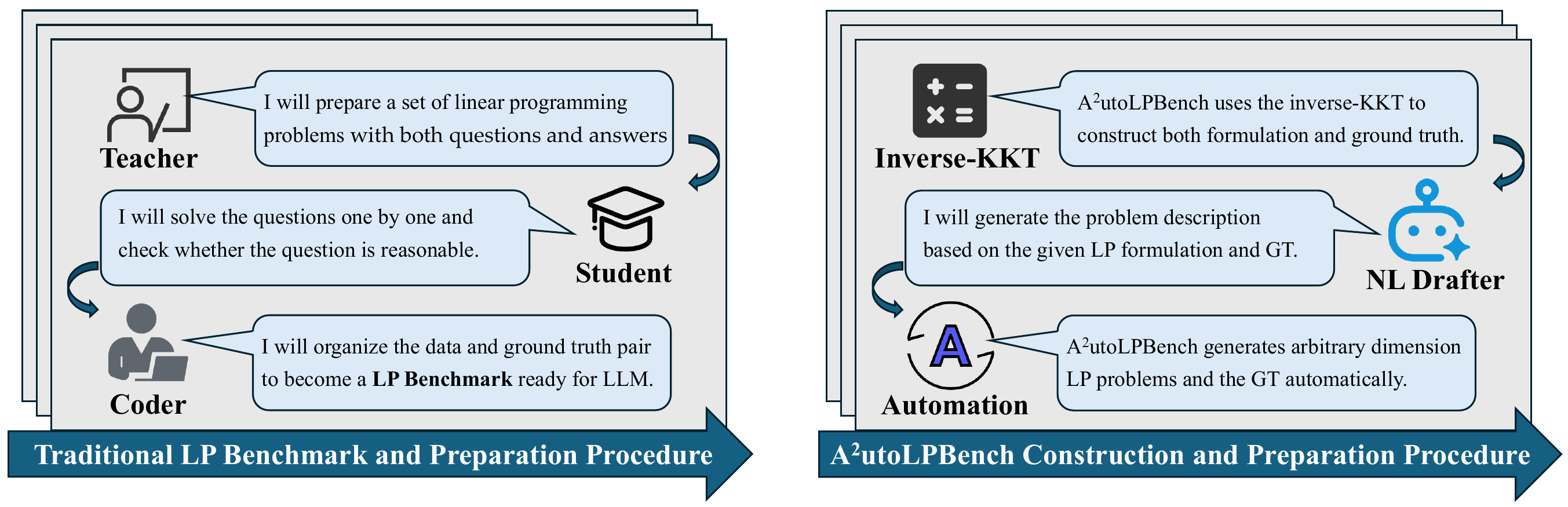}
    \caption{Construction pipelines compared and the motivation of the A$^{2}$utoLPBench. Left: traditional LP benchmarks require three coordinated human roles (Teacher / Student / Coder). Right: A$^{2}$utoLPBench replaces them with an inverse-KKT routine, an LLM-driven Natural Language (NL) drafter, and an automation layer that streams instances at any seed and dimension, removing per-item human labor and making the supply and the difficulty distribution programmable.}
    \label{fig:intro_pipeline}
\end{figure}

We address all four limitations by shipping a \textbf{generator} instead of a corpus, together with an \textbf{agent-runnable evaluation surface}. \Cref{fig:intro_pipeline} contrasts the traditional human-in-the-loop construction with our automated pipeline. \textbf{A$^{2}$utoLPBench} has two parts, Auto and Agent.
The Auto part is the inverse-KKT generator: given dimensions $(n, m)$ and a random seed, it draws a primal-dual-coefficient triple $(x, \lambda, A)$ and reverse-derives the LP $(c=A^{\top}\lambda,\, b=Ax)$ whose optimum is, by KKT, exactly $\phi = c^{\top} x$, with no solver invocation and no human annotator in the loop.
The Agent part is the bundled evaluation surface: a solver-critic baseline (\Cref{sec:method-dualagent}) and a Docker environment with agent-readable documentation (\Cref{app:docker}) whose declared tool-use skills any tool-use-capable agent can register and run plug-and-play.
The two parts are designed together. The generator addresses limitations (i)--(iii) by producing fresh, parametrically-tunable, contamination-free items on demand. The agent surface addresses limitation (iv) by giving downstream agents a single \texttt{docker run} entry point.
Re-framing the LP benchmark as a generator rather than a corpus yields structural properties that fixed benchmarks cannot have. These properties follow from the generator-versus-corpus design choice rather than from any empirical difficulty claim. \Cref{tab:properties} summarizes them. Our main contributions are summarized as follows:

\begin{enumerate}
    \item Unlimited supply at low LLM-side cost: the generator produces LP instances in closed form from sampled triples $(x,\lambda,A)$ via $b=Ax$ and $c=A^{\top}\lambda$, plus a single LLM drafter call. Per-item cost is dominated by that drafter call rather than human author or annotator labor (cf.\ NL4OPT~\cite{ramamonjison2022nl4opt} and MAMO~\cite{huang2024mamo}), so benchmark size decouples from annotator budget. We ship a $256$-instance reference snapshot; the procedure scales to any size and seed range (\Cref{sec:dataset}).

    \item Parametric, reproducible difficulty: difficulty is set by the generator's $(n,m)$ knobs and coefficient ranges, not fixed at release time. The generator traces a sol-rate curve over eight strata from $(2,3)$ to $(40,40)$ on which existing benchmarks appear as isolated points; independently sampled batches with the same parameters agree to within $3$pp off-cliff, turning the benchmark from a single score into a measurement instrument, illustrated by~\Cref{fig:scaling_envelope} in~\Cref{app:scaling_envelope}).

    \item Mathematically certified ground truth: for each instance, the optimum $\phi=c^{\top}x$ is established by the KKT theorem (\Cref{thm:optimality}) before any solver is invoked, removing both solver-precision drift and human-annotator transcription error. The certificate covers the algebraic LP $(A,b,c)$; the natural-language rendering $T$ is checked by a string-level coefficient match, with the limits of single-pass rendering discussed in \Cref{app:nl_limits}.

    \item Contamination resistance by construction: any batch drawn from a seed range chosen after a model's training cutoff cannot have appeared in that model's training corpus. The released snapshot can itself leak into future training data once published, but downstream evaluators can always draw fresh seed ranges to obtain contamination-free batches; the guarantee is structural, tied to the generator-vs-corpus distinction, rather than an empirical claim about any particular file.
\end{enumerate}

\begin{table}[t!]
    \centering
    \caption{Structural properties of A$^{2}$utoLPBench versus existing fixed-corpus LP benchmarks.}
    \label{tab:properties}
    \small
    \begin{tabular}{l l l l}
        \toprule
        & \textbf{NL4OPT}\footnotemark & \textbf{MAMO}\footnotemark & \textbf{A$^2$utoLPBench} \\
        \midrule
        Item supply             & 287       & 211 / 652\footnotemark   & $\infty$         \\
        Parametric difficulty   & one level & easy/complex& tunable via $(n,m)$       \\
        GT source               & human     & human       & KKT theorem      \\
        Construction cost       & human-authored & human-authored & one LLM drafter call \\
        Cross-batch reproducible& N/A       & N/A         & $\le 3$pp off-cliff        \\
        Contamination resistant & vulnerable after release & vulnerable after release & yes for fresh seed ranges \\
        \bottomrule
    \end{tabular}
\end{table}
\addtocounter{footnote}{-3}%
\stepcounter{footnote}\footnotetext{NL4OPT release: \url{https://github.com/nl4opt/nl4opt-competition}.}%
\stepcounter{footnote}\footnotetext{MAMO paper: \url{https://arxiv.org/abs/2405.13144v1}.}%
\stepcounter{footnote}\footnotetext{MAMO ships two subsets: $211$ complex problems and $652$ easy problems. Both subsets are used in our experimental comparison (\Cref{sec:exp}).}%

In addition to the generator itself, the Agent part of A$^{2}$utoLPBench provides two pieces of agent-side tooling for plug-and-play deployment.
The first is a dual-agent solver-critic baseline (\Cref{sec:method-dualagent}) organized around three actions, Propose, Audit, and Refine, that on external LP benchmarks where models score well below ceiling recovers up to $+15.1$pp on Kimi-K2.5 and $+11.3$pp on Claude-Sonnet-4.6 over MAMO complex through a bounded-iteration loop.
The second is an agent-friendly Docker environment (\Cref{app:docker}) that bundles the generator, verifier, dual-agent runtime, and an \texttt{AGENTS.md} usage manual targeted at LLM-driven agents, so that a downstream agent can attach to the image and obtain a calibrated LP score with a single \texttt{docker run}.

\paragraph{Related work.}
Existing LP-from-text benchmarks~\cite{ramamonjison2022nl4opt,yang2024optibench,ahmaditeshnizi2023optimus,ahmaditeshnizi2024optimus} are static corpora in the sense of~\Cref{eq:benchmark_def}: $\mathcal{D}$ is fixed at release time and each $\phi_i$ is obtained from a solver run or a human annotator, exposing them to training-corpus contamination~\cite{mirzadeh2024gsm,xu2024benchmark} and a frozen difficulty distribution as models improve~\cite{wang2025benchmark}. Generator-based reasoning benchmarks~\cite{zhu2023dyval,fan2024nphardeval} address contamination by parameterising instance generation; A$^{2}$utoLPBench extends that idea to LP-from-text and pairs it with a closed-form KKT certificate (\Cref{thm:optimality}) that also removes solver dependence. The bundled solver-critic protocol of \Cref{sec:method-dualagent} is closely related to self-refinement~\cite{madaan2023self,shinn2023reflexion,gou2023critic}, sharing the documented caveat that internal self-correction can degrade on already-saturated tasks~\cite{huang2023large}. \Cref{app:related-work} expands these comparisons and adds the verifiable-reasoning, OR-LLM agent, and tool-use-benchmark literatures.

\section{Preliminaries}
\label{sec:prelim}

\subsection{Linear programming and KKT optimality}
\label{sec:prelim:lp}

A linear program (LP) optimizes a linear objective subject to linear inequality and non-negativity constraints. We consider the standard maximization form
\begin{equation}
    \label{eq:lp}
    \begin{aligned}
        \max_{x} \quad & c^\top x, \\
        \text{s.t.} \quad       & Ax \le b, \\
                                & x \ge 0,
    \end{aligned}
\end{equation}
where $c \in \mathbb{R}^n$, $A \in \mathbb{R}^{m \times n}$, and $b \in \mathbb{R}^m$. The optimal objective value is denoted $\phi = \max_{x} c^{\top}x$ subject to the constraints in~\Cref{eq:lp}, and a maximizer is denoted $x^{\star}$. Every primal LP of the form~\Cref{eq:lp} admits a dual problem $\min_{\lambda \ge 0} b^{\top}\lambda$ subject to $A^{\top}\lambda \ge c$, and strong duality holds whenever both primal and dual are feasible.

\paragraph{KKT optimality.}
For LPs, the Karush--Kuhn--Tucker (KKT) conditions are both necessary and sufficient for optimality: a feasible point $x$ achieves the optimum of~\Cref{eq:lp} if and only if there exist dual variables $\lambda\in\mathbb{R}_{\ge 0}^{m}$ and $\mu\in\mathbb{R}_{\ge 0}^{n}$ such that
\begin{subequations}\label{eq:kkt}
\begin{align}
    Ax &\le b, \qquad x\ge 0, \label{eq:kkt:primal}\\
    A^\top \lambda - \mu &= c, \label{eq:kkt:stationarity}\\
    \lambda_i\bigl(b_i-(Ax)_i\bigr) &= 0 \ \ \forall i,\qquad \mu_j x_j = 0 \ \ \forall j. \label{eq:kkt:cs}
\end{align}
\end{subequations}
\Cref{eq:kkt:primal} is primal feasibility, \Cref{eq:kkt:stationarity} is stationarity (equivalent to dual feasibility under our sign convention on $\mu$), and \Cref{eq:kkt:cs} is complementary slackness on the inequality constraints and non-negativity bounds, respectively. These conditions provide the algebraic identities that we exploit in \Cref{sec:method-generation} to construct LP instances directly from a sampled optimum, rather than solving for one.

\subsection{Agent formalization for LP-from-text}
\label{sec:prelim:agent}

We model an LP-from-text agent $\mathcal{A}$ as a stochastic mapping that takes a natural-language description $T$ and produces a numerical answer $\hat{z}\in\mathbb{R}$:
\begin{equation}
    \label{eq:agent_def}
    \mathcal{A}: T \;\longmapsto\; \hat{z}.
\end{equation}
$T$ describes an underlying LP $(A, b, c)$ in the form of~\Cref{eq:lp}, with every coefficient of $A$, $b$, and $c$ stated explicitly in the text and the optimization sense (maximize/minimize) and variable bounds named in plain language. Internally an agent typically realizes $\mathcal{A}$ as a multi-stage pipeline,
\begin{equation}
    \label{eq:agent_pipeline}
    T
    \;\xrightarrow{f_{\text{parse}}}\;
    F
    \;\xrightarrow{f_{\text{code}}}\;
    \pi
    \;\xrightarrow{f_{\text{exec}}}\;
    \hat{z},
\end{equation}
where $F = (\widehat{A}, \widehat{b}, \widehat{c}, \text{sense}, \text{bounds})$ is the formulation extracted from $T$, $\pi$ is an executable solver program (typically Python that calls \texttt{scipy.optimize.linprog} or \texttt{pulp}), and $\hat{z}$ is the final printed objective value. In this paper, we treat $\mathcal{A}$ as a black box and score only $\hat{z}$; \Cref{sec:method-dualagent} reuses the decomposition~\Cref{eq:agent_pipeline} to characterize the failure modes the bundled solver-critic protocol is designed to catch.

\paragraph{Benchmark and score.}
An \textbf{LP-from-text instance} is a pair $(T,\phi)$ where $\phi$ is the ground-truth optimum of the LP underlying $T$, and a \textbf{benchmark} is a finite collection $\mathcal{D}=\{(T_i,\phi_i)\}_{i=1}^{N}$. Existing LP-from-text benchmarks are \textbf{static-corpus benchmarks}: $\mathcal{D}$ is fixed at release time and each $\phi_i$ is obtained from a solver invocation or a human annotation, so $|\mathcal{D}|$ is bounded by the annotator budget. We instead construct a \textbf{generator-based benchmark}, a randomized procedure $\mathcal{G}: s \mapsto (T,\phi)$ that produces fresh instances on demand from a random seed $s$ and yields $\mathcal{D}_{\mathcal{G}}(S)=\{\mathcal{G}(s):s\in S\}$ on any chosen seed range $S$, with $|S|$ controllable at evaluation time and each $\phi$ established by the KKT theorem rather than by a solver run (\Cref{sec:method-generation}). Formal definitions and the equations underlying these notations are given in \Cref{app:formalization}.

For an agent $\mathcal{A}$ on a benchmark $\mathcal{D}$ we report the sol-rate
\begin{equation}
    \label{eq:solrate}
    S(\mathcal{A}, \mathcal{D}) = \frac{1}{|\mathcal{D}|}\sum_{i=1}^{|\mathcal{D}|} \mathbb{1}\bigl[\,\text{correct}(\hat{z}_i, \phi_i)\,\bigr],
\end{equation}
where $\hat{z}_i = \mathcal{A}(T_i)$ and $\text{correct}(\hat{z}, \phi)$ is a relative-error judgment whose tolerance is fixed throughout the paper and stated in \Cref{sec:exp:setup}. The notation $\mathcal{A}$, $T$, $\hat{z}$, $\phi$, $\mathcal{D}$, $\mathcal{D}_{\mathcal{G}}(S)$, $S(\cdot,\cdot)$ is used consistently from this point on.

\section{Method}
\label{sec:method}

\begin{figure}[t!]
    \centering
    \includegraphics[width=0.78\linewidth]{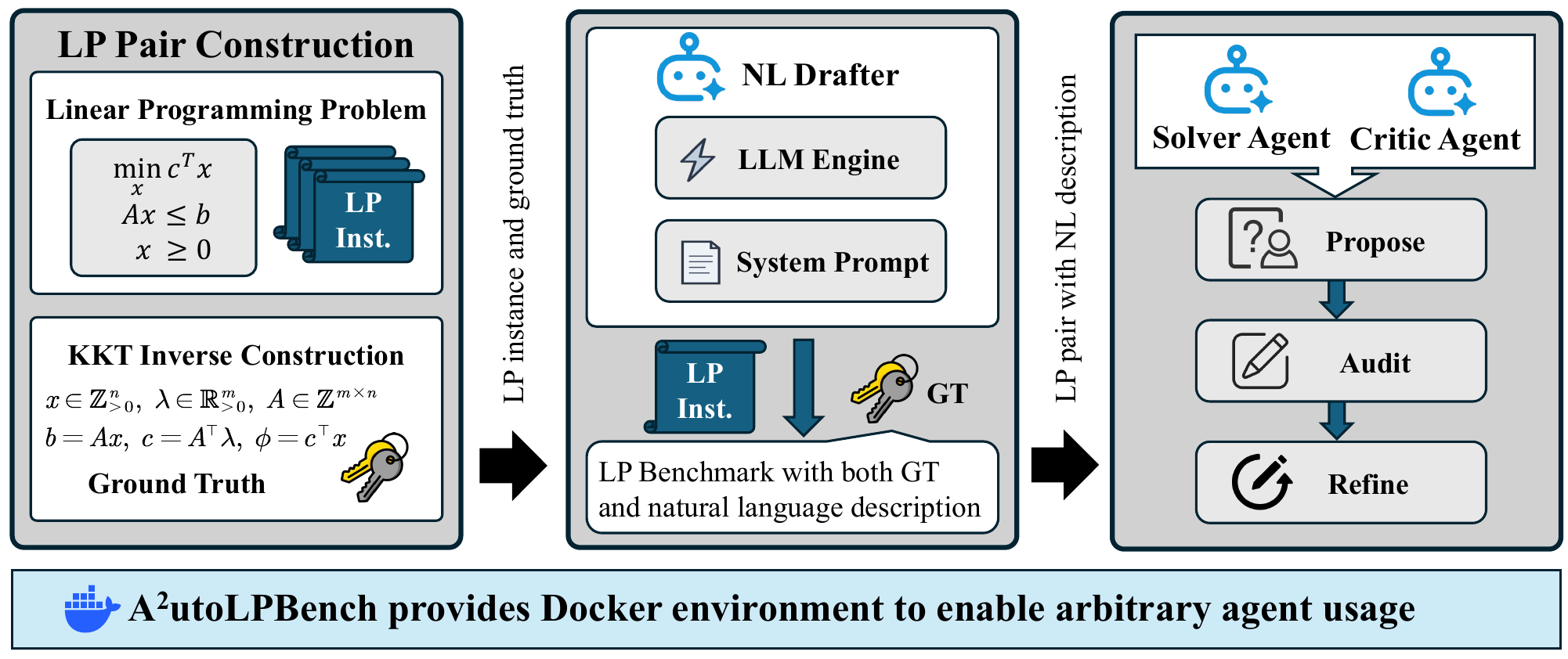}
    \caption{Overview of A$^{2}$utoLPBench. (i) inverse-KKT constructs an LP pair, (ii) an LLM drafter renders it to a natural-language specification, (iii) at inference time the dual-agent solver-critic protocol Propose / Audit / Refine produces and iteratively revises candidate solutions.}
    \label{fig:pipeline}
\end{figure}

This section presents the four components of A$^{2}$utoLPBench: the inverse-KKT generator (\Cref{sec:method-generation}, the Auto part), the bundled solver-critic baseline and the agent-runnable Docker runtime (\Cref{sec:method-dualagent} and \Cref{sec:method-docker}, the two halves of the Agent part), and the released 256-instance reference snapshot (\Cref{sec:dataset}). \Cref{fig:pipeline} gives an end-to-end overview. Three LLM-driven agents appear across these components: the natural-language drafter that converts each LP triple into a word problem (system and user prompts in \Cref{app:nlp_prompt}), the solver agent that proposes a candidate Python program (system prompt in \Cref{app:solver-prompt}), and the critic agent that audits the candidate and decides whether to refine (system prompt in \Cref{app:critic-prompt}). Notation follows \Cref{sec:prelim}.

\paragraph{What makes LP-from-text hard.}
Recall the agent pipeline of~\Cref{eq:agent_pipeline}: $T \to F \to \pi \to \hat{z}$. For the final $\hat{z}$ to be correct, three independent capabilities must hold simultaneously. (i) The formulation $F$ extracted by $f_{\text{parse}}$ must agree with the LP underlying $T$. (ii) The program $\pi$ produced by $f_{\text{code}}$ must respect the solver's calling convention (maximize-vs-minimize sign, row-vs-column matrix layout, default variable bounds). (iii) The executed value $\hat{z} = f_{\text{exec}}(\pi)$ must survive both the solver's internal precision and the agent's output formatting. The typical failure modes correspond directly to violations of these three: $\widehat{A}$ transposed row-by-row instead of column-by-column, a ``$\le$'' constraint silently recorded as ``$\ge$'', a maximization passed to a minimization solver without coefficient negation, a single constraint dropped from $\widehat{b}$, or $\hat{z}$ overflowing when $n$ and $m$ are large. \Cref{app:case-study} traces representative instances of each failure mode, and \Cref{app:scaling_envelope} shows that the cumulative failure rate climbs sharply as $(n,m)$ grow. The four components below address these failure modes from complementary angles: a generator with mathematically certified ground truth (\Cref{sec:method-generation}), an inference-time solver-critic runtime that audits and refines candidate outputs (\Cref{sec:method-dualagent}), a Docker environment that lets any LLM-driven agent run the benchmark end-to-end (\Cref{sec:method-docker}), and a released reference dataset (\Cref{sec:dataset}).

\subsection{Auto Part: LP Pair Construction via Inverse-KKT}
\label{sec:method-generation}

We refer to $(A, b, c)$ as the LP instance, $(x, \phi)$ as its ground truth (GT), and the combination of the LP instance and its ground truth as an LP pair, which is $(A, b, c, x, \phi)$.

A central challenge in scalable benchmark construction is ground-truth reliability: pipelines that sample random instances and certify the optimum with an external solver inherit solver dependence and floating-point fragility, while pipelines that rely on human annotation inherit annotation errors and cannot be cheaply regenerated.
A$^{2}$utoLPBench instead constructs instances from the optimality conditions themselves, focusing on candidate points $x > 0$ that we will certify optimal by KKT. At such an interior point, complementary slackness on the non-negativity constraints forces $\mu = 0$, so the KKT system of \Cref{sec:prelim} reduces to the two linear equations $A^{\top}\lambda = c$ and $Ax = b$, together with strict positivity $x > 0$ and $\lambda > 0$.
We sample $x \in \mathbb{Z}_{>0}^{n}$, a full-rank $A \in \mathbb{Z}^{m\times n}$ with $\operatorname{rank}(A) = \min(m,n)$ (a numerical-diversity guard described in \Cref{app:reserved}; not required for the optimality proof), and strictly positive $\lambda \in \mathbb{R}_{>0}^{m}$ (Algorithm~\ref{alg:inverse}), and set
\begin{equation}
    \label{eq:inverse-def}
    b = A x,
    \qquad
    c = A^{\top} \lambda,
\end{equation}
and record the ground-truth optimum
\begin{equation}
    \label{eq:invkkt_optval}
    \phi = c^{\top} x.
\end{equation}
By construction, all primal constraints are tight at $x$ and dual feasibility holds with equality, which is exactly what is needed for the KKT optimality conditions.

By construction, $(A, b, c, x)$ with $x > 0$ and $\lambda > 0$ satisfies every KKT condition of the LP in \Cref{eq:lp} (taking the multiplier $\mu = 0$ on the non-negativity block), so $x$ is an optimal solution with optimal value $\phi = c^{\top} x$. The formal statement and its full block-by-block KKT verification are given as \Cref{thm:optimality} in \Cref{app:proof}.

The generator exposes three knobs as parameters of \Cref{alg:inverse}: $(n, m)$, the coefficient ranges of $A$ and $x$, and the scaling of $\lambda$. \Cref{app:scaling_envelope} sweeps the $(n, m)$ axis while holding the other two fixed at the defaults of \Cref{app:reserved}, tracing out the difficulty curve we report there. The full pseudocode is given as \Cref{alg:inverse} in \Cref{app:reserved}.

\paragraph{Natural-language rendering.}
Each LP pair is converted into a natural-language word problem $T$ via a single-shot LLM drafter (the prompt is given in \Cref{app:nlp_prompt}).
The drafter is instructed to use every coefficient verbatim, name the optimization direction, and state every constraint explicitly; we verify numerical fidelity by string-matching every coefficient against the rendered text and resampling if any coefficient is missing.
This single-pass procedure suffices for the difficulty regimes evaluated in this paper; we discuss the limits of single-pass NL rendering in \Cref{app:nl_limits}.

\subsection{Agent Part: Bundled LP Solver}
\label{sec:method-dualagent}

A static dataset can surface the failure modes characterized at the start of \Cref{sec:method} only after the fact. The Agent part of A$^{2}$utoLPBench instead provides an inference-time runtime that audits candidate outputs against those failure modes and refines them on the fly, shipped as a bundled solver-critic baseline alongside the dataset, so that any downstream evaluation has a stronger reference point than vanilla LLM zero-shot inference.

The protocol involves two agents and three actions. The solver agent $\mathcal{A}$ (system prompt in \Cref{app:solver-prompt}) takes the natural-language specification $T$ and produces an executable Python program whose final printed line is the candidate optimum $\hat{z}$. The critic agent $\mathcal{C}$ (system prompt in \Cref{app:critic-prompt}) takes $T$, the candidate program, and its standard output, and emits a verdict $V \in \{\textsc{Agree}, \textsc{Disagree}\}$ together with a short rationale. Each round of the loop consists of one or more of three actions.

\paragraph{Propose.}
The solver $\mathcal{A}$ reads $T$ and produces an initial candidate $y_{1}$. This action runs once per item and corresponds, in isolation, to vanilla LLM zero-shot inference.

\paragraph{Audit.}
The critic $\mathcal{C}$ reads $(T, y_{k})$ and emits $(V_{k}, r_{k})$: a binary verdict and a textual rationale. This action runs once per round.

\paragraph{Refine.}
If the audit returns \textsc{Disagree} and the budget is not exhausted, the solver re-enters the loop conditioned on the rationale $r_{k}$ and produces a new candidate $y_{k+1}$. This action runs at most $K-1$ times per item.

\smallskip
\noindent
With budget $K \in \mathbb{N}_{>0}$, the round-$k$ outputs are
\begin{equation}
    \label{eq:dualagent_loop}
    \begin{aligned}
        y_{1} \;&\sim\; p_{\mathcal{A}}(\cdot\mid T)
        && \text{(Propose)}, \\[2pt]
        (V_{k}, r_{k}) \;&\sim\; p_{\mathcal{C}}(\cdot\mid T, y_{k})
        && \text{(Audit)}, \\[2pt]
        y_{k+1} \;&\sim\; p_{\mathcal{A}}(\cdot\mid T, r_{k})
        && \text{(Refine, if } V_{k} = \textsc{Disagree}\text{ and } k < K\text{)}.
    \end{aligned}
\end{equation}
The loop terminates at the smallest $k$ with $V_{k} = \textsc{Agree}$, or at $k = K$.
An optional LP-modeling skill, a curated tutorial that names canonical failure modes, can be prepended to the solver's system prompt; we treat skill and critic as orthogonal axes.
Full system prompts and the coverage analysis $P_{K} \ge 1 - (1 - \pi_{1})^{K}$ are deferred to \Cref{app:dualagent}.
\Cref{sec:exp:transfer} reports its empirical effect: a meaningful uplift on benchmarks where models are well below ceiling (notably MAMO complex), and near-zero uplift on already-saturated benchmarks.

\subsection{Agent Part: Docker Runtime}
\label{sec:method-docker}

Beyond the solver-critic baseline of \Cref{sec:method-dualagent}, the Agent part of A$^{2}$utoLPBench ships an agent-runnable evaluation surface as a self-contained Docker image (\texttt{autolpbench:latest}) that preloads the inverse-KKT generator, KKT verifier, scipy/PuLP solvers, and the bundled solver-critic runtime. Inside the image, an \texttt{AGENTS.md} manual declares four tool-use skills (problem fetch, solver invocation, KKT verification, batch generation), each described in the JSON-schema format consumed by current tool-use APIs (OpenAI function calling, Anthropic tools, etc.). Any LLM-driven agent that supports tool use can read \texttt{AGENTS.md}, register the declared skills, and run the benchmark plug-and-play, without further glue code or environment setup.

This packaging closes the four static-corpus limitations (i)--(iv) listed in \Cref{sec:intro}: the bundled generator addresses (i)--(iii), and the \texttt{AGENTS.md} manifest addresses (iv) by giving downstream agents a single \texttt{docker run} entry point. Full image specification, the tool-use skill schema, and a sample \texttt{AGENTS.md} excerpt are in \Cref{app:docker}.

\subsection{Dataset Construction in Practice}
\label{sec:dataset}

A$^{2}$utoLPBench is a generator, not a corpus: we ship a 256-instance reference snapshot stratified into eight size classes $(n,m)$ from $(2,3)$ to $(40,40)$ at 32 instances per class, but the same generator produces additional instances at any size and seed, so a downstream evaluator can draw a fresh contamination-free batch at any time (per-stratum seed ranges, JSON schema, and per-item wall-time in \Cref{app:reserved}). Each instance is produced by the inverse-KKT procedure of \Cref{alg:inverse} followed by a single LLM-drafter call. Ground truth is correct by construction via \Cref{thm:optimality} and does not depend on a solver; as a redundant sanity check we also re-run scipy's \texttt{linprog} on every released $(A, b, c)$ and confirm that its numerical solution matches $\phi$ within relative tolerance $\tau=10^{-4}$ (all 256 pass). For comparison, NL4OPT~\cite{ramamonjison2022nl4opt} releases 287 human-authored word problems and MAMO~\cite{huang2024mamo} releases 211 complex + 652 easy items, both as fixed crowdsourced corpora. The released snapshot, the generator, and the solver-critic runtime are bundled in the Docker image of \Cref{sec:method-docker}.

\section{Experiments}
\label{sec:exp}
\subsection{Experimental Setup}
\label{sec:exp:setup}
The experiments come in two pieces. \textbf{(i) Calibration}: we sweep DeepSeek-V4 across all eight strata of A$^{2}$utoLPBench in three independent seed ranges, then re-run three additional solvers (MiMo-V2.5, Qwen3.5, Kimi-K2.5) on the same instances. To flip the drafter side, that is, to vary the LLM that authors the natural-language description $T$ in \Cref{sec:method-generation}, we additionally evaluate all four solvers on a batch in which DeepSeek-V4 (rather than the default MiMo-V2.5) drafts the descriptions; we refer to these two batches as ``MiMo-drafted'' and ``DS-drafted'' below. The per-stratum DeepSeek-V4 sol-rate runs from $100\%$ on the small strata to $8.3\%$ at s40x40 with batch-to-batch noise $\le 3$pp off the cliff, supporting the parametric-difficulty and cross-batch claims of \Cref{tab:properties}. Full curve, four-solver figure, and cross-distribution findings are in \Cref{app:scaling_envelope}. \textbf{(ii) External-benchmark transfer}: we evaluate the bundled solver-critic baseline on NL4OPT and the two MAMO subsets under DeepSeek-V4 / MiMo-V2.5 cross-pairings, then extend to a six-solver leaderboard with a fixed MiMo-V2.5 critic; this is also where token-cost auditing lives (\Cref{sec:exp:transfer}).

Every solver runs in two inference modes: \textbf{vanilla} (direct generation of solver code, no critic) and \textbf{A$^{2}$utoLPBench} (the bundled solver-critic baseline of \Cref{sec:method-dualagent}, configured at $K=3$ critic rounds). All models are queried at temperature $0.3$ for reproducibility; reasoning-capable solvers (Qwen3.5, Kimi-K2.5, OpenAI-GPT5.4, Claude-Sonnet-4.6) run with thinking / extended-reasoning modes disabled to match the deployment regime that downstream agent users typically face. The executor sandbox imports \texttt{scipy.optimize.linprog} and \texttt{pulp} and verifies output against $\phi$ via the relative-error criterion $|\phi - \hat{z}|/(|\phi| + 10^{-8}) < 10^{-2}$. Per-item logs and aggregate summaries are released alongside the dataset.

\subsection{External-Benchmark Transfer}
\label{sec:exp:transfer}

We further evaluate A$^{2}$utoLPBench's bundled solver-critic baseline on two external static-corpus benchmarks. Across each of NL4OPT and the two MAMO subsets (easy and complex), we use a fixed prefix of $53$ items per cell, the largest equal-size subset that fits the cross-product of solver-critic configurations within our compute and rate-limit budget; using a fixed prefix rather than a random subsample keeps comparisons reproducible across cells, and we acknowledge the fixed-prefix choice as a sampling limitation that a random-subsample sensitivity check would tighten. With this subset fixed, we pair two solver families (DeepSeek-V4 and MiMo-V2.5) under four configurations: vanilla, A$^{2}$utoLPBench with the same-model critic, and A$^{2}$utoLPBench with each cross-model critic combination. \Cref{tab:bigtable} reports per-cell sol-rate.

\begin{table}[t!]
    \centering
    \caption{External-benchmark transfer: sol-rate (\%) of two solver families under vanilla and A$^{2}$utoLPBench inference on the NL4OPT, MAMO easy, and MAMO complex subsets defined in \Cref{sec:exp:transfer}. ``A$^{2}$LPBench $X$+$Y$'' denotes solver $X$ paired with critic $Y$ in the bundled $K=3$ solver-critic loop. The same abbreviation is used in subsequent tables.}
    \label{tab:bigtable}
    \small
    \setlength{\tabcolsep}{5pt}
    \begin{tabular}{l c c c}
        \toprule
        \textbf{Configuration} & \textbf{NL4OPT} & \textbf{MAMO easy} & \textbf{MAMO complex} \\
        \midrule
        Vanilla DeepSeek-V4                       & 98.11 & 67.92 & 26.42 \\
        Vanilla MiMo-V2.5                     & 98.11 & 79.25 & 13.21 \\
        \midrule
        A$^{2}$LPBench DeepSeek-V4+DeepSeek-V4 (same-model)     & 98.11 & 71.70 & 33.96 \\
        A$^{2}$LBenchP MiMo-V2.5+MiMo-V2.5 (same-model) & 90.57 & 83.02 & 24.53 \\
        A$^{2}$LPBench DeepSeek-V4+MiMo-V2.5 (cross-model)  & 98.11 & 73.58 & 30.19 \\
        A$^{2}$LPBench MiMo-V2.5+DeepSeek-V4 (cross-model)  & 100.00 & 81.13 & 28.30 \\
        \bottomrule
    \end{tabular}
\end{table}

Three patterns appear consistently. (a) On NL4OPT, vanilla DeepSeek-V4 and MiMo-V2.5 are already saturated at $98.11\%$, leaving no room for the solver-critic loop to add value; the only configuration that meaningfully deviates is MiMo-V2.5+MiMo-V2.5, which loses $7.5$pp because a same-distribution critic occasionally over-disagrees on already-correct code. (b) On MAMO complex, the underlying benchmark has substantial headroom (vanilla $13.2$--$26.4\%$); every A$^{2}$utoLPBench configuration yields a positive delta, with the largest uplift coming from cross-model pairings of the weaker solver: MiMo-V2.5+DeepSeek-V4 reaches $28.3\%$ from a $13.2\%$ vanilla baseline ($+15.1$pp). (c) On MAMO easy, gains are smaller and direction-dependent.

Comparing this transfer pattern to the calibration sweep, where same-model A$^{2}$LPBench (abbreviated as A$^{2}$LP) DeepSeek-V4+DeepSeek-V4 lifts the s40x40 sol-rate from $8.3\%$ to $33.3\%$ ($+25$pp; numerical bound check in \Cref{app:dualagent}, full curve in \Cref{app:scaling_envelope}), we observe that the preferred critic-pairing depends on which failure mode dominates. When failures are dominated by formulation errors (\textit{e.g.}\ MAMO complex), a cross-model critic catches errors the solver's own distribution does not flag. When failures are dominated by numerical precision (\textit{e.g.}\ A$^{2}$utoLPBench s40x40), same-model critics produced fewer harmful disagreements on numerically delicate instances; \Cref{app:case-study} gives trace-level evidence. The bundled solver-critic baseline therefore generalizes across benchmarks but its preferred critic-pairing is benchmark-dependent.

\paragraph{Multi-solver leaderboard with a fixed cross-model critic.}
To check that the cross-model A$^{2}$utoLPBench protocol generalizes beyond DeepSeek-V4 and MiMo-V2.5, we run vanilla and ``$X$ + MiMo-V2.5 critic'' configurations across six solver families on the same MAMO complex subset. \Cref{tab:leaderboard_xmodel} reports per-cell sol-rate.

\begin{table}[!htb]
    \centering
    \caption{Multi-solver leaderboard on MAMO complex. All A$^{2}$utoLPBench cells use MiMo-V2.5 as the critic, so the MiMo-V2.5 + MiMo-V2.5 same-model entry is N/A here (the same-model self-loop appears in \Cref{tab:bigtable} as a separate same-distribution diagnostic). OpenAI-GPT5.4 is queried via the OpenAI Chat Completions API with \texttt{reasoning\_effort=none}.}
    \label{tab:leaderboard_xmodel}
    \small
    \begin{tabular}{l c c c}
        \toprule
        \textbf{Solver} & \textbf{Vanilla} & \textbf{A$^{2}$LPBench $X$+MiMo-V2.5} & \textbf{$\Delta$} \\
        \midrule
        DeepSeek-V4         & 26.42 & 30.19 & $+3.77$ \\
        MiMo-V2.5           & 13.21 & N/A   & N/A \\
        Qwen3.5             & 20.75 & 28.30 & $+7.55$ \\
        Kimi-K2.5           & 15.09 & 30.19 & $+15.10$ \\
        OpenAI-GPT5.4       & 15.09 & 13.21 & $-1.89$ \\
        Claude-Sonnet-4.6   & 20.75 & 32.08 & $+11.32$ \\
        \bottomrule
    \end{tabular}
\end{table}

Of the five solvers paired with a cross-model MiMo-V2.5 critic (MiMo-V2.5 itself is excluded since it is the critic), four gain, with the largest uplift on Kimi-K2.5 ($+15.1$pp) and Claude-Sonnet-4.6 ($+11.3$pp). The sole regression among these five is OpenAI-GPT5.4 ($-1.9$pp). Inspecting its A$^{2}$utoLPBench traces shows that the critic disagreed on $14$ items at the Audit stage, but the solver successfully Refined only $2$ of them (about $14\%$); on the remaining $12$ items the Refined output either repeated the same error or executed an unrelated formulation that produced an empty stdout. The traces suggest that, under \texttt{reasoning\_effort=none} in this configuration, OpenAI-GPT5.4 Proposes a single LP formulation efficiently but does not robustly incorporate critic feedback into a corrective Refine step. The bundled solver-critic loop therefore acts as a capability amplifier that requires the underlying solver to act on critic feedback at the Refine stage; when the solver is below that threshold, the loop does not help and can occasionally regress on items vanilla would have solved. The full failure analysis with traces is in \Cref{app:gpt-regression}. We treat this as a limitation of the bundled solver-critic baseline rather than a property of the benchmark, and recommend reporting both vanilla and A$^{2}$utoLPBench scores by default (as we do throughout \Cref{tab:leaderboard_xmodel}) so that the loop's contribution is visible per solver. The structural properties of \Cref{tab:properties} (parametric difficulty, KKT-certified ground truth, contamination resistance) follow from the inverse-KKT generator (\Cref{sec:method-generation}) and hold for any inference protocol, regardless of whether the loop adds or subtracts on a particular solver.

\paragraph{Inference-time token cost.}
We log every API call's prompt and completion token counts directly from the OpenAI / Anthropic SDK \texttt{usage} field, then aggregate to per-item totals. \Cref{tab:tokens} reports per-item mean tokens (summed across all iterations $k = 1, \dots, K$) on the same MAMO complex subset for the configurations of \Cref{tab:leaderboard_xmodel}. The aggregate uses every cell's actual \texttt{dc\_avg\_iter}, so the totals are directly comparable across solvers without normalising to a fixed $k$.

\begin{table}[!htb]
    \centering
    \caption{Per-item token usage (mean across all iterations) on the complex subset of MAMO\cite{huang2024mamo} benchmark. ``solver $p/c$'' and ``critic $p/c$'' are aggregated prompt / completion tokens summed over the $K \le 3$ iterations actually executed. The $\sim 1{,}200$-token gap between vanilla and A$^{2}$utoLPBench solver prompts is the LP-optimization specification prepended to the system prompt. Per-item monetary cost is auditable from these counts and the publicly listed prices on each provider's official site at the time of writing; the surrounding paragraph reports two reference points (DeepSeek-V4 list rate and Anthropic Claude-Sonnet-4.6 list rate).}
    \label{tab:tokens}
    \small
    \setlength{\tabcolsep}{3pt}
    \begin{tabular}{l c c c c c c}
        \toprule
        Configuration & avg.\ iters & solver $p$ & solver $c$ & critic $p$ & critic $c$ & avg. tokens \\
        \midrule
        Vanilla DeepSeek-V4                  & 1.00 &    685 &    925 & N/A   & N/A &  1{,}610 \\
        Vanilla MiMo-V2.5                & 1.00 &    689 &    554 & N/A   & N/A &  1{,}243 \\
        Vanilla Qwen3.5             & 1.00 &    883 & 2{,}264 & N/A   & N/A &  3{,}147 \\
        Vanilla Kimi-K2.5           & 1.00 &    817 & 1{,}285 & N/A   & N/A &  2{,}101 \\
        Vanilla OpenAI-GPT5.4             & 1.00 &    816 &    483 & N/A   & N/A &  1{,}298 \\
        Vanilla Claude-Sonnet-4.6         & 1.00 &    834 &    742 & N/A   & N/A &  1{,}576 \\
        \midrule
        A$^{2}$LPBench DeepSeek-V4+MiMo-V2.5           & 1.30 & 2{,}450 & 1{,}450 & 1{,}477 & 1{,}091 &  6{,}468 \\
        A$^{2}$LPBench Qwen3.5+MiMo-V2.5      & 1.45 & 4{,}059 & 4{,}517 & 2{,}569 & 1{,}001 & 12{,}146 \\
        A$^{2}$LPBench Kimi-K2.5+MiMo-V2.5    & 1.40 & 2{,}816 & 3{,}230 & 2{,}828 &     48 &  8{,}922 \\
        A$^{2}$LPBench OpenAI-GPT5.4+MiMo-V2.5      & 1.40 & 2{,}842 &    824 & 2{,}085 &     76 &  5{,}828 \\
        A$^{2}$LPBench Claude-Sonnet-4.6+MiMo-V2.5  & 1.36 & 2{,}881 & 1{,}332 & 2{,}462 &     62 &  6{,}737 \\
        \bottomrule
    \end{tabular}
\end{table}

\paragraph{Pricing the loop and the dataset.}
Vanilla cells run at $1.2$--$3.1$K tokens per item and A$^{2}$utoLPBench cells at $5.8$--$12.1$K tokens per item, so the solver-critic loop costs roughly $3$--$5\times$ a single vanilla call. We monetise the token totals from \Cref{tab:tokens} at the publicly listed per-token list prices on each provider's official site (May 2026): one A$^{2}$LP DeepSeek-V4+MiMo-V2.5 item costs $\sim \$0.0023$ at DeepSeek-V4 list pricing, and one A$^{2}$LP Claude-Sonnet-4.6+MiMo-V2.5 item costs $\sim \$0.030$ at Anthropic's Claude-Sonnet-4.6 rate. 

\textbf{Dataset construction is cheaper still}: each new instance costs one LLM-drafter call, with median per-item cost $\sim\$0.0015$ at DeepSeek-V4 list rate and $\sim\$0.019$ at Claude-Sonnet-4.6 list rate. The savings come from removing the human author and annotator from the loop entirely; full token breakdowns, per-stratum drafter cost, and the largest-instance prices are in \Cref{app:cost-detail}.

\section{Conclusion}
\label{sec:conclusion}

In this paper, we presented \textbf{A$^{2}$utoLPBench}, an LP-from-text benchmark that pairs an inverse-KKT generator (KKT-certified ground truth, parametric difficulty across eight strata, contamination-resistant under fresh seed ranges) with an agent-runnable solver-critic baseline and a Docker image. The bundled loop transfers to external LP benchmarks with up to $+15.1$pp uplift on Kimi-K2.5 over MAMO complex; together, the two parts move LP evaluation from static corpora toward calibrated, contamination-resistant generators.

\bibliographystyle{plainnat}
\bibliography{ref/Top-sim,ref/reference}

\newpage
\appendix

\section{Auto-Part Implementation Details}
\label{app:reserved}

\Cref{alg:inverse} below describes the high-level construction; the main-text discussion (\Cref{sec:method-generation}) summarizes the same procedure in prose. This appendix documents the concrete sampling distributions, rank safeguards, coefficient ranges, the natural-language drafter prompt, and the per-stratum composition of the 256-instance reference snapshot used in all experiments of \Cref{sec:exp}. The same procedure scales to arbitrary instance counts: downstream evaluators can call the generator with new seed ranges to obtain fresh, contamination-free batches at any size.

\begin{algorithm}[htbp]
  \caption{Inverse LP Construction with Ground-Truth Guarantee}
  \label{alg:inverse}
  \begin{algorithmic}[1]
    \REQUIRE Dimensions $n$ (variables), $m$ (constraints); coefficient ranges $[a_{\min}, a_{\max}]$, $[x_{\min}, x_{\max}]$, $[\lambda_{\min}, \lambda_{\max}]$; rank-attempt budget $K_{\text{rank}}$
    \ENSURE LP pair $(A, b, c, x, \lambda, \phi)$ certified optimal by KKT

    \vspace{2pt}
    \STATE \textbf{// Phase 1: Sample primal, coefficients (with full-rank guard), dual}
    \STATE $x \gets \textsc{SamplePrimal}(n;\, x_{\min}, x_{\max})$ \hfill \COMMENT{positive integers, ensures $x > 0$}
    \FOR{$k = 1, 2, \ldots, K_{\text{rank}}$}
        \STATE $A \gets \textsc{SampleCoefficients}(m, n;\, a_{\min}, a_{\max})$
        \IF{$\operatorname{rank}(A) = \min(m, n)$}
            \STATE \textbf{break} \hfill \COMMENT{full rank; numerical-diversity guard}
        \ENDIF
    \ENDFOR
    \IF{$\operatorname{rank}(A) < \min(m, n)$}
        \STATE $A \gets \textsc{IdentityBackboneFallback}(m, n;\, a_{\min}, a_{\max})$ \hfill \COMMENT{deterministic full-rank fallback}
    \ENDIF
    \STATE $\lambda \gets \textsc{SampleDual}(m;\, \lambda_{\min}, \lambda_{\max})$ \hfill \COMMENT{strictly positive multipliers}

    \vspace{2pt}
    \STATE \textbf{// Phase 2: Derive the LP and the ground truth in closed form}
    \STATE $b \gets A x$ \hfill \COMMENT{makes every primal constraint active at $x$}
    \STATE $c \gets A^{\!\top}\! \lambda$ \hfill \COMMENT{dual feasibility holds with equality on $x>0$}
    \STATE $\phi \gets c^{\!\top}\! x$ \hfill \COMMENT{optimal value certified by~\Cref{thm:optimality}}

    \vspace{2pt}
    \STATE \textbf{return} $(A, b, c, x, \lambda, \phi)$
  \end{algorithmic}
\end{algorithm}

\begin{remark}[Why Tight Constraints]
  Setting $b = Ax$ (line~14) makes every primal constraint active at the optimum.
  Complementary slackness $\lambda_i (b_i - (Ax)_i) = 0$ then holds with $b_i - (Ax)_i = 0$, so any strictly positive $\lambda$ is admissible without violating KKT.
  Together with $c = A^{\top}\lambda$ this gives a closed-form pair $(x, \lambda)$ that satisfies all KKT conditions for~\Cref{eq:lp}, certifying $\phi = c^{\top} x$ as the optimum without invoking a solver.
\end{remark}

\subsection{Proof of the Ground-Truth Guarantee}
\label{app:proof}

\begin{theorem}[Ground-truth guarantee]
    \label{thm:optimality}
    Let $(A, b, c, x)$ be constructed as in~\Cref{eq:inverse-def} with $x > 0$ and $\lambda > 0$. Then $x$ is an optimal solution of the LP in~\Cref{eq:lp} with optimal value $\phi = c^{\top} x$.
\end{theorem}

\begin{proof}
The KKT system of \Cref{eq:kkt} for an LP requires four blocks of conditions: primal feasibility, dual feasibility, stationarity, and complementary slackness. Take $\mu = 0 \in \mathbb{R}^n$ as the multiplier on the non-negativity constraint $x \ge 0$, and verify each block in turn.

\paragraph{Primal feasibility.}
Since $x > 0$ entrywise we have $x \ge 0$. The construction sets $b = Ax$, so $Ax = b$ and the inequality $Ax \le b$ holds with equality. Both constraints in \Cref{eq:kkt:primal} are therefore satisfied.

\paragraph{Dual feasibility.}
The construction provides $\lambda > 0$, so $\lambda \in \mathbb{R}_{\ge 0}^{m}$ as required. We chose $\mu = 0 \in \mathbb{R}_{\ge 0}^{n}$.

\paragraph{Stationarity.}
The stationarity condition \Cref{eq:kkt:stationarity} reads $A^{\top}\lambda - \mu = c$. With $\mu = 0$ this reduces to $A^{\top}\lambda = c$, which is the second equation of the inverse-KKT construction in \Cref{eq:inverse-def}.

\paragraph{Complementary slackness.}
The inequality complementary-slackness condition $\lambda_i (b_i - (Ax)_i) = 0$ holds for every $i$ because $b - Ax = 0$ by construction (the parenthetical factor is identically zero). The non-negativity complementary-slackness condition $\mu_j x_j = 0$ holds for every $j$ because $\mu = 0$.

\paragraph{Sufficiency for LP optimality.}
A linear program is a convex program (its objective and constraints are linear, hence both convex and concave), and its feasible set $\{x : Ax \le b, x \ge 0\}$ is a polyhedron, for which the linearity constraint qualification holds whenever the feasible set is non-empty. Standard convex-optimization theory~\citep{boyd2004convex} therefore guarantees that the four KKT blocks verified above are sufficient for global optimality, so $x \in \arg\max\{c^{\top} x' : A x' \le b,\, x' \ge 0\}$. The corresponding optimal value is
\begin{equation*}
    \phi \;=\; \max_{x' \ge 0,\, A x' \le b} c^{\top} x' \;=\; c^{\top} x,
\end{equation*}
matching the recorded ground truth in \Cref{eq:invkkt_optval}.
\end{proof}

\paragraph{Remark on uniqueness.}
The theorem certifies that $x$ is one optimal solution, not the unique one: LPs can admit multiple optima when the objective is parallel to a face of the feasible polytope. We claim no uniqueness of the maximizer in the construction, only of the optimal value $\phi$, which is unique under strong duality whenever an optimum exists.

\paragraph{Remark on the interior assumption.}
The construction restricts to interior optima $x > 0$ so that $\mu = 0$ is admissible by inspection. For boundary optima with $x_j = 0$ for some $j$, the corresponding $\mu_j$ would need to be chosen to satisfy stationarity, complicating the closed-form derivation. Restricting to $x > 0$ does not meaningfully narrow the structural diversity of the LP family, since every LP with a finite optimum can be reformulated with positive offsets into an equivalent interior-optimum instance.

\subsection{Reference Snapshot Composition}

\Cref{tab:dataset_stats} records the composition of the initial $256$-instance batch we ship as a convenience reference. The release size $256$ is a starting point for cross-paper comparison, not a cap on dataset size: the generator produces additional batches of any size, dimension, and seed range on demand, and the per-stratum item counts shown below can be increased arbitrarily without further engineering.

\begin{table}[!htb]
    \centering
    \caption{Composition of the initial $256$-instance batch shipped with the release. The $256$ figure is illustrative only; the generator produces additional instances of any size, dimension, and seed range on demand, so the per-stratum counts below should be read as the configuration we used in \Cref{sec:exp}, not as the upper bound of the dataset.}
    \label{tab:dataset_stats}
    \small
    \begin{tabular}{l c c r}
        \toprule
        Stratum & $(n,m)$ & Seed range & \# items \\
        \midrule
        s2x3        & $(2,3)$    & 300\,001--300\,032 & 32 \\
        s3x4        & $(3,4)$    & 300\,033--300\,064 & 32 \\
        s5x6        & $(5,6)$    & 300\,065--300\,096 & 32 \\
        s10x12      & $(10,12)$  & 300\,097--300\,128 & 32 \\
        s15x15      & $(15,15)$  & 300\,129--300\,160 & 32 \\
        s20x20      & $(20,20)$  & 300\,161--300\,192 & 32 \\
        s30x30      & $(30,30)$  & 300\,193--300\,224 & 32 \\
        s40x40      & $(40,40)$  & 300\,225--300\,256 & 32 \\
        \midrule
        \textbf{Total} &  &  & \textbf{256} \\
        \bottomrule
    \end{tabular}
\end{table}

\subsection{Inverse-KKT Sampling}

\paragraph{Primal solution $x$.}
Each entry $x_j$ is drawn independently and uniformly from $\{1,2,3,4,5\}$ (positive integers).
Integer values keep the natural-language description readable and ensure $x>0$ so the interior-optimum condition (\Cref{sec:prelim}) holds without further checks.

\paragraph{Constraint matrix $A$.}
Entries $a_{ij}$ are drawn independently and uniformly from $\{1,2,3,4,5\}$ (positive integers, no zeros).
We then verify $\operatorname{rank}(A)=\min(m,n)$; if not, we resample up to $100$ attempts.
After $100$ rejections we fall back to a deterministic full-rank backbone (an identity-like pattern with random non-zero filler) so the generator never blocks indefinitely.
In practice we have not observed the fallback firing for any $(n,m)$ in $\{(2,3),\ldots,(40,40)\}$.

\paragraph{Dual multipliers $\lambda$.}
Entries $\lambda_i$ are drawn independently and uniformly from $[0.5, 3.0]$ (real-valued, strictly positive).
Setting $b = Ax$ in the next step makes every primal inequality active at $x$, after which any strictly positive $\lambda$ satisfies complementary slackness $\lambda_i (b_i - (Ax)_i) = 0$ trivially (the parenthetical is zero); we use $\lambda > 0$ rather than $\lambda \ge 0$ for numerical-diversity reasons rather than for the proof.

\paragraph{Derived quantities.}
Given $(x, A, \lambda)$, we set $b = Ax$, $c = A^{\top}\lambda$, $\phi = c^{\top} x$.
By construction $b$ is integer-valued and $c$ is real-valued (a positive linear combination of integer columns of $A$ weighted by real $\lambda$).
We do not round $c$ or $\phi$; the natural-language drafter is asked to use every coefficient verbatim (\Cref{app:nlp_prompt}).

\paragraph{Per-trial artifact format.}
Each instance is serialized as one JSON line with the following keys: \texttt{instance\_id}, \texttt{split}, \texttt{n}, \texttt{m}, \texttt{lp\_form}, \texttt{A}, \texttt{b}, \texttt{c}, \texttt{ground\_truth}~$\{$\texttt{x\_star}, \texttt{phi\_star}, \texttt{lambda\_star}, \texttt{active\_set}$\}$, \texttt{nl\_description}, \texttt{generation}~$\{$\texttt{seed}, \texttt{drafter\_model}, \texttt{all\_constraints\_tight}, \texttt{lambda\_positive}$\}$, and \texttt{verification}~$\{$\texttt{tau}, \texttt{eps}, \texttt{verified}$\}$.
This schema mirrors the 256-instance reference snapshot at our anonymous repository.

\paragraph{Construction throughput.}
The full pipeline (numerical sampling, KKT verification, NL drafting, and fidelity check) takes approximately $0.1$~s of compute per item, excluding LLM-drafter latency, which is dominated by network IO and amortizes to $\sim 1$~s per item under standard parallelism. Producing the 256-instance reference snapshot took approximately 12 minutes of wall time on a single workstation; the corresponding NL4OPT and MAMO collection efforts are crowdsourced human-authored corpora and require human-author plus annotator labor per item~\cite{ramamonjison2022nl4opt}.

\paragraph{Drafter token usage and per-item cost.}
\label{app:cost-detail}
\Cref{sec:exp:transfer} prices the solver-critic loop and the dataset-construction call at headline rates. We provide the underlying token counts here. Counting the drafter prompt and the rendered description with the \texttt{tiktoken} \texttt{gpt-4o} encoding over all 256 v1.0 instances, the per-item drafter cost has a median of $1.6$K input plus $0.9$K output tokens, ranging from $0.7$K total at $(2,3)$ to $18$K total at $(40,40)$ as the matrix grows. Pricing the median item gives $\sim\$0.0015$ at DeepSeek-V4 list rate and $\sim\$0.019$ at Claude-Sonnet-4.6 list rate; the largest $(40,40)$ items reach $\sim\$0.012$ and $\sim\$0.16$ respectively. Producing fresh batches at scale is therefore bounded by drafter latency rather than monetary cost. We report only LLM-side costs and avoid putting a specific number on annotator labor since published per-item annotation rates are not consistently reported across NL4OPT and MAMO.

\subsection{Natural-Language Drafter}
\label{app:nlp_prompt}

The drafter is a single-shot LLM call that receives the LP triple $(A,b,c)$ together with sense and bounds and emits a self-contained natural-language word problem. We use temperature $0.6$ and the prompts below.

\paragraph{System prompt.}\par\smallskip\noindent
\begin{promptbox}
You are an expert at converting mathematical optimization problems into vivid and concrete natural language word problems.
You can transform an abstract linear program into a classic OR scenario such as resource allocation, production planning, or diet optimization.
Your descriptions are accurate, vivid, and logically consistent.
\end{promptbox}

\paragraph{User prompt template.}
The user message embeds the LP coefficients and demands numerical fidelity:\par\smallskip\noindent
\begin{promptbox}
Convert the following Linear Programming problem into a complete natural-language problem description.

Linear Programming Problem ($n$ variables, $m$ constraints): \\
- Objective coefficients (maximize): $c$ \\
- Constraint coefficient matrix $A$: \\
- Constraint RHS vector: $b$ \\
- Constraint directions: $\{\leq\}^m$ \\
- Variable bounds: $x \geq 0$

Strict requirements:
\textbf{(1) Numerical fidelity is mandatory.} Every coefficient in $c$, every entry of $A$, every RHS in $b$ must appear explicitly in your description, in the exact same numerical value. Do not round, do not abbreviate, do not say ``and so on'', do not summarize ranges.
\textbf{(2)} For larger problems you may use compact in-line list notation, e.g.\ ``the profit per unit for the $n$ products is $[v_1, v_2, \ldots, v_n]$''.
\textbf{(3)} Wrap the lists in a real-world scenario so the description reads as a problem statement, not a raw matrix dump.
\textbf{(4)} The narrative must correspond exactly to the given numerical structure.
\textbf{(5)} Use English. \textbf{(6)} Variables are continuous.
\end{promptbox}

\paragraph{Fidelity check.}
After drafting, we string-match every coefficient in $A$, $b$, $c$ against the rendered text.
If any coefficient is missing, we resample the drafter at the same temperature.
On the 256-instance reference snapshot, the first-pass fidelity rate is $100\%$ ($256/256$); no resampling was triggered.

\section{Benchmark Formalization}
\label{app:formalization}

This appendix expands the benchmark notation summarized in \Cref{sec:prelim:agent}. The two design patterns we distinguish, \textbf{static-corpus} and \textbf{generator-based}, differ in how $\mathcal{D}$ is sourced and in how each $\phi_i$ is established.

\paragraph{Items, ground truth, and benchmarks.}
An \textbf{LP-from-text instance} is a pair $(T, \phi)$ in which $T$ is the natural-language description from~\Cref{eq:agent_def} and $\phi$ is the ground-truth optimum of the underlying LP in~\Cref{eq:lp}. A \textbf{benchmark} is a finite collection
\begin{equation}
    \label{eq:benchmark_def}
    \mathcal{D} \;=\; \{(T_i, \phi_i)\}_{i=1}^{N},
    \qquad |\mathcal{D}| = N,
\end{equation}
together with the implied LP coefficients $(A_i, b_i, c_i)$ that underlie each $T_i$. Two design choices distinguish benchmarks: how $\mathcal{D}$ is sourced, and how each $\phi_i$ is established.

\paragraph{Static-corpus and generator-based benchmarks.}
A static-corpus benchmark fixes $\mathcal{D}$ at release time as a distributed file; the size $|\mathcal{D}|$ is bounded by the annotator budget invested during curation, and each $\phi_i$ is typically obtained from a solver invocation or a human annotation. A generator-based benchmark replaces the fixed file with a generator $\mathcal{G}\colon s \mapsto (T, \phi)$ that produces fresh instances on demand from a random seed $s$; the realized benchmark on a chosen seed range $S$ is
\begin{equation}
    \label{eq:generator_benchmark}
    \mathcal{D}_{\mathcal{G}}(S) \;=\; \{\,\mathcal{G}(s) : s \in S\,\},
    \qquad |\mathcal{D}_{\mathcal{G}}(S)| = |S|,
\end{equation}
with $|S|$ controllable at evaluation time. We construct one such $\mathcal{G}$ in \Cref{sec:method-generation} where each $\phi$ is established directly by the KKT theorem without solver invocation.

\section{Limits of Single-Pass Natural-Language Rendering}
\label{app:nl_limits}

The single-pass drafter is sufficient for the difficulty regimes evaluated in this paper.
Two limitations are worth flagging.

\textbf{Layout-level difficulty.}
Different drafter models produce different $(A,b)$ layouts in prose.
Two common patterns are (i) a separated b-vector at the end of the description (``the total available capacity is $[b_1, \ldots, b_m]$'') and (ii) inline RHS interleaved with each row (``Resource $i$ usage: $[\ldots]$, availability $= b_i$'').
At larger $(n,m)$ the inline pattern is materially harder for a downstream solver agent to parse correctly.
A more careful drafter that controls layout explicitly, or a specification of NL style as part of the dataset card, would let evaluators isolate this from problem-size difficulty.

\textbf{Single-domain scenarios.}
The current drafter prompt biases toward generic OR scenarios (production planning, resource allocation, diet).
Domain-specific styles (e.g., supply-chain logistics with named locations, energy dispatch with grid topology) are not represented in the release.
Specializing the drafter prompt is straightforward but is left to future work.

\section{Dual-Agent Solver-Critic: Full Specification}
\label{app:dualagent}

\Cref{sec:method-dualagent} gives the high-level loop of \Cref{eq:dualagent_loop}.
Here we provide the full system prompts of the solver agent (\Cref{app:solver-prompt}) and the critic agent (\Cref{app:critic-prompt}), together with the coverage analysis. The third agent in our pipeline, the natural-language drafter, has its system and user prompts in \Cref{app:nlp_prompt}.

\needspace{8\baselineskip}
\subsection{Solver Agent: System Prompt}
\label{app:solver-prompt}
\noindent
\begin{promptbox}
You are a Python programming expert skilled in using scipy for optimization problems.
Your task is to generate Python code using \texttt{scipy.optimize.linprog} to solve linear programming problems.

Requirements:
(1) Code must be complete and executable.
(2) Use \texttt{scipy.optimize.linprog} for solving.
(3) Print the solution: optimal variables, objective value.
(4) Code must be enclosed in a Python code block.
(5) Note that \texttt{linprog} minimizes by default; negate coefficients for maximization.

\textbf{OUTPUT CONTRACT (REQUIRED).} The very last line your code prints must be exactly:
\verb|FINAL_ANSWER: <number>|.
\end{promptbox}

\subsection{Critic Agent: System Prompt}
\label{app:critic-prompt}
\noindent
\begin{promptbox}
You are a meticulous reviewer for LP solutions.
You will receive: the original problem statement, a candidate Python solution (source + stdout).
Decide whether the formulation and answer are faithful.
Look for: (1) matrix transposition; (2) wrong objective direction; (3) missing or extra constraints; (4) wrong constraint signs; (5) variable-bound errors; (6) numerical scale anomalies.

Output format (strict):
\verb|VERDICT: AGREE|
or
\verb|VERDICT: DISAGREE|
\verb|REASON: <one or two sentences>|.

If unsure, prefer \texttt{AGREE} over speculative \texttt{DISAGREE}; false alarms waste solver iterations.
\end{promptbox}

\paragraph{LP-modeling primer prepended to the solver prompt.}
In addition to the solver and critic prompts above, we prepend a short curated primer to the solver's system prompt by default. The primer names canonical failure modes (matrix transposition, sign errors, missing bounds), shows a worked example, and demonstrates the \verb|FINAL_ANSWER:| output contract. The primer is part of the bundled A$^{2}$utoLPBench inference mode and is enabled in all experiments unless stated otherwise; ablations isolating its effect are out of scope for this paper.

\paragraph{Coverage analysis.}
Let $\mathcal{S}$ denote the success region (candidates within the verifier's relative-error tolerance), and let
$\pi_{1} = \Pr_{y \sim p_{\mathcal{A}}(\cdot \mid T)}[\,y \in \mathcal{S}\,]$ be the solver's single-shot success rate.
We adopt three assumptions on the dual-agent protocol.
(i) \textbf{Critic faithfulness}: when $y_k \in \mathcal{S}$, $\Pr[V_k = \textsc{Agree} \mid y_k] = 1$.
(ii) \textbf{Critic conservatism}: when $y_k \notin \mathcal{S}$, $\Pr[V_k = \textsc{Disagree} \mid y_k] = 1$.
(iii) \textbf{Round non-degradation}: conditional on rounds $1,\ldots,k-1$ all failing, $\pi_k \geq \pi_1$.

Under these assumptions, the $K$-round dual-agent protocol satisfies
\begin{equation}
P_K = \Pr[y_{\mathrm{final}} \in \mathcal{S}] \;\geq\; 1 - \prod_{k=1}^{K}(1-\pi_k) \;\geq\; 1-(1-\pi_1)^K.
\label{eq:coverage_app}
\end{equation}
The proof follows from (i) and (ii) jointly: $V_k = \textsc{Agree}$ holds if and only if $y_k \in \mathcal{S}$, so the verdict is a faithful indicator of correctness. The loop therefore terminates at round $k$ with $y_{\mathrm{final}} \in \mathcal{S}$ exactly when $y_k$ is the first round-$k$ candidate inside $\mathcal{S}$, and $y_{\mathrm{final}} \notin \mathcal{S}$ holds if and only if every round produces $y_k \notin \mathcal{S}$. Under (iii) the joint failure probability factorizes into per-round factors each at most $1-\pi_1$, yielding $\Pr[y_{\mathrm{final}} \notin \mathcal{S}] \le (1-\pi_1)^K$.

\paragraph{Empirical check against the bound.}
We anchor the bound at two regimes from the main experiments.

\textbf{Low-saturation regime.} On AutoLPBench s40x40 with DeepSeek-V4 (\Cref{app:scaling_envelope}), we measure $\pi_1 = 0.083$, so $K=3$ gives a theoretical lower bound of $1-(1-0.083)^3 \approx 0.229$. The observed dual-agent success rate is $0.333$, which satisfies the bound with $10.4$pp of margin and is consistent with the picture that the bundled solver-critic loop has substantial room to lift when the underlying solver is well below ceiling.

\textbf{High-saturation, same-distribution regime.} On MAMO complex with MiMo-V2.5 paired same-model (\Cref{tab:bigtable}), we measure $\pi_1 = 0.132$, so $K=3$ gives a bound of $0.346$, and the observed dual-agent success rate is $0.245$, $10.1$pp below the bound. The bound is therefore not satisfied here. Both assumptions (i) and (ii) of \Cref{eq:coverage_app} are meaningfully strained when the solver and critic share a distribution. Assumption (i) (faithfulness) fails when the critic disagrees on candidates that are already within tolerance, and these false-negative disagreements pull the realized success rate down; assumption (ii) (conservatism) fails when the critic agrees on candidates that are not within tolerance, and these false-positive agreements terminate the loop with an incorrect $y_{\mathrm{final}}$. \Cref{app:case-study}, item \texttt{mamo\_complex\_43}, contains a concrete trace of the false-positive failure mode (the critic returns \textsc{Agree} on an empty stdout). The combined effect is consistent with the $7.5$pp regression of A$^{2}$LP MiMo-V2.5+MiMo-V2.5 versus vanilla MiMo-V2.5 on saturated NL4OPT in \Cref{tab:bigtable}.

We accordingly treat \Cref{eq:coverage_app} as a design heuristic for choosing $K$ given a target $\pi_1$ and a deployment regime, and recommend reporting the realized $P_K$ alongside the bound for any deployment rather than substituting the bound for the realized number.

\paragraph{Verdict-stratified output distribution.}
A second consequence of the critic-faithfulness assumption is that the final verdict $V$ stratifies outputs: cases ending in \textsc{Agree} have markedly higher success rate than those that reach the budget cap with \textsc{Disagree}.
This calibration signal is reported as a runtime confidence indicator and is consistent with the trace-level evidence in \Cref{app:case-study}.

\section{Parametric Difficulty Envelope: Full Curve}
\label{app:scaling_envelope}

\Cref{sec:exp} of the main text reports the headline calibration numbers; here we provide the full per-stratum curve and the cross-distribution analysis across four solvers.

\begin{figure}[!htb]
    \centering
    \includegraphics[width=0.78\linewidth]{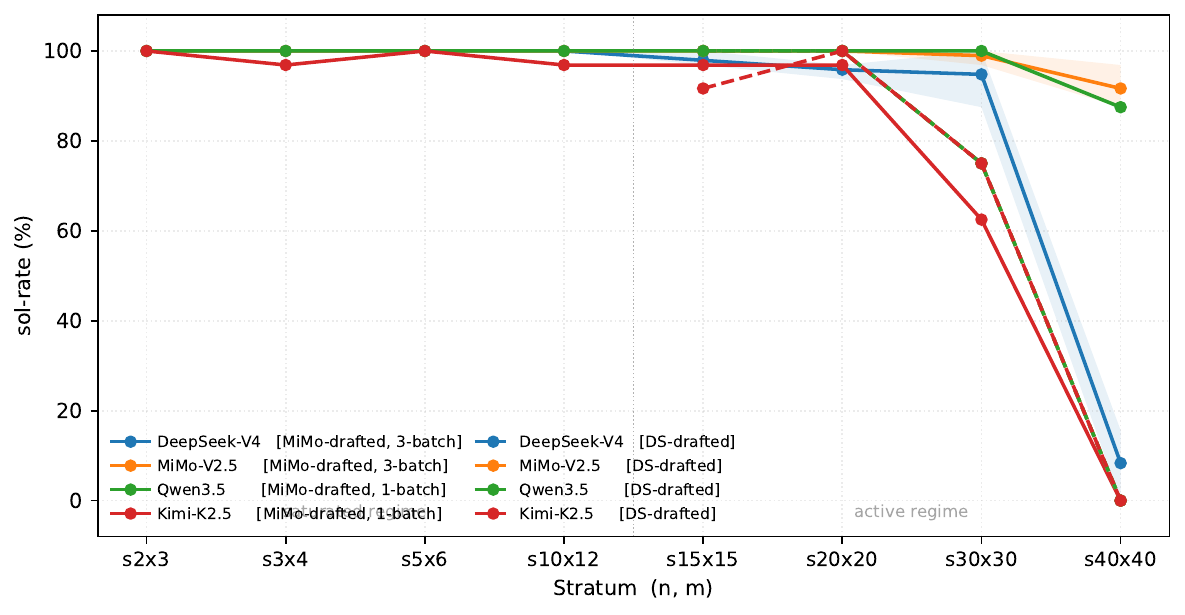}
    \caption{Per-stratum vanilla sol-rate of four solvers across the eight strata. Color = solver (DeepSeek-V4: blue, MiMo-V2.5: orange, Qwen3.5: green, Kimi-K2.5: red). \textbf{Solid lines}: instances drafted by MiMo-V2.5 (3-batch max--min envelope shown for DeepSeek-V4 and MiMo-V2.5; Qwen3.5 and Kimi-K2.5 are single-batch point estimates). \textbf{Dashed lines}: instances drafted by DeepSeek-V4 (single batch \texttt{xl\_47} v2, big strata only). A$^{2}$utoLPBench surfaces a difficulty cliff at s40x40 that fixed-corpus LP benchmarks cannot expose; on the cliff stratum the four solvers split into two clusters under MiMo-drafting (MiMo-V2.5 + Qwen3.5 vs DeepSeek-V4 + Kimi-K2.5), and collapse to a uniform $0\%$ under DS-drafting. Per-stratum readings and the four-observation interpretation follow below.}
    \label{fig:scaling_envelope}
\end{figure}

Four observations follow from the curves.

First, on DeepSeek-V4 the sol-rate is saturated at $100\%$ for the four small strata (s2x3 through s10x12), degrades to $\sim 95\%$ through the mid range (s15x15--s30x30), and falls to $8.3\%$ at s40x40, an $\sim 87$pp cliff between adjacent strata. This is the kind of capability boundary that fixed benchmarks cannot surface: NL4OPT corresponds to the saturated easy end and MAMO complex sits in the mid range for current SOTA models, but no public LP benchmark contains items at the s40x40 size class.

Second, the 3-batch envelope widths for DeepSeek-V4 and MiMo-V2.5 are $0$pp at saturation, $\le 3$pp through the mid range, and widen to $12$--$16$pp around the s30x30--s40x40 cliff. This shape tracks the binomial standard deviation $\sigma_{\hat{p}}=\sqrt{p(1-p)/32}$ and is therefore a property of the per-stratum sample size we chose to release, not an artifact of any particular seed range.

Third, on the cliff stratum (s40x40, MiMo-drafted) the four solvers split into two clusters: MiMo-V2.5 ($91.7\%$) and Qwen3.5 ($87.5\%$) versus DeepSeek-V4 ($8.3\%$) and Kimi-K2.5 ($0\%$). The split does not track simple drafter-equals-solver alignment, since Qwen3.5 (which did not draft these instances) is in the high cluster while Kimi-K2.5 is in the low cluster. One hypothesis consistent with the split is deployment-stack or prompt/tokenization proximity between the drafter and the high-cluster solvers (MiMo-V2.5 drafted these instances, and Qwen3.5 shares the Aliyun deployment stack with MiMo-V2.5). We do not attempt to formalize this from a single experiment, and we cannot rule out alternative explanations including training-data-family proximity, tokenizer convention, drafter-style memorization, or prompt-template artifact; the observation is reported as evidence that solver identity at the difficulty cliff matters and that the cliff is not a fixed property of $(n,m)$.

Fourth, on the DS-drafted batch the four solvers are essentially indistinguishable at every stratum. All four reach $100\%$ at s15x15 and s20x20, all four sit at $75\%$ at s30x30, and all four collapse to $0\%$ at s40x40. Drafter style at large $(n,m)$ therefore swamps solver identity. Combined with the third observation, this gives us two complementary mechanisms by which the s40x40 cliff manifests, both of which a fixed-corpus benchmark cannot probe because it cannot vary either the drafter or the seed range.

\section{Case Study: Critic-Driven Rescues}
\label{app:case-study}

To understand why and when the bundled solver-critic loop helps, we examine the items on which a vanilla solver produces an incorrect answer but the A$^{2}$utoLPBench loop with the same solver-base recovers the correct answer (a ``rescue''), and the converse (a ``regression''). \Cref{tab:rescue-counts} summarizes the rescue/regression counts on the MAMO complex subset across the four A$^{2}$utoLPBench configurations from \Cref{tab:bigtable}.

\begin{table}[!htb]
    \centering
    \caption{Per-cell rescue and regression counts on MAMO complex. A rescue is an item where vanilla is wrong and A$^{2}$utoLPBench is correct; a regression is the converse.}
    \label{tab:rescue-counts}
    \small
    \begin{tabular}{l c c c}
        \toprule
        Vanilla $\rightarrow$ A$^{2}$utoLPBench & Rescues & Regressions & Net \\
        \midrule
        DeepSeek-V4 $\rightarrow$ DeepSeek-V4+DeepSeek-V4         &  4 & 0 & $+4$ \\
        DeepSeek-V4 $\rightarrow$ DeepSeek-V4+MiMo-V2.5 (cross)   &  4 & 2 & $+2$ \\
        MiMo-V2.5 $\rightarrow$ MiMo-V2.5+MiMo-V2.5               &  6 & 0 & $+6$ \\
        MiMo-V2.5 $\rightarrow$ MiMo-V2.5+DeepSeek-V4 (cross)     &  8 & 0 & $+8$ \\
        \bottomrule
    \end{tabular}
\end{table}

Two patterns are visible. First, all four configurations have non-negative net rescue: the bundled loop never makes more items wrong than it fixes. Second, on the weaker solver baseline (MiMo-V2.5, where vanilla sol-rate is $13.2\%$), pairing with a different-distribution critic (DeepSeek-V4) yields the largest net rescue ($+8$, with zero regressions); on the stronger solver baseline (DeepSeek-V4, where vanilla sol-rate is $26.4\%$), the cross-model critic introduces some regressions ($-2$), and the same-model critic gives the cleanest net result.

\paragraph{Qualitative walkthroughs.}
Below we trace the critic's intervention on four representative rescues, using the MiMo-V2.5 $\rightarrow$ MiMo-V2.5+DeepSeek-V4 configuration (which gave the largest net rescue) and the DeepSeek-V4 $\rightarrow$ DeepSeek-V4+DeepSeek-V4 configuration (which yielded clean same-model rescues).

\begin{enumerate}
    \item \textbf{Empty output (item \texttt{mamo\_complex\_46}):} the vanilla MiMo-V2.5 solver returned a code block whose execution produced no \texttt{FINAL\_ANSWER} line; the candidate stdout was empty. The DeepSeek-V4 critic flagged this on iteration~$1$ (``the candidate code produced no output, so no final answer is provided''), the solver re-derived the formulation, and the second iteration printed $\phi = 4813$, matching the certified ground truth.
    \item \textbf{Missing constraint (item \texttt{mamo\_complex\_23}):} the problem statement listed two protein requirements ($88$g and $144$g, modeled as a maximum constraint plus a minimum constraint). The vanilla MiMo-V2.5 solver retained only the binding constraint and dropped the other. The DeepSeek-V4 critic disagreed twice with increasingly precise feedback (``the candidate enforces only one protein constraint and adds an unnecessary zero variable bound''), and the solver's third attempt produced $\phi \approx 23.84$ (within the $1\%$ tolerance of the certified $\phi = 24$).
    \item \textbf{Wrong constraint type (item \texttt{mamo\_complex\_41}):} the vanilla MiMo-V2.5 solver modeled a flow-balance constraint as $\sum_{out} - \sum_{in} = \text{net}$ when the problem allowed slack on either side. The DeepSeek-V4 critic flagged that ``total surplus ($674$) exceeds total deficit ($398$), making strict equality infeasible''; the solver switched to inequality constraints and produced $\phi = 2114$.
    \item \textbf{Semantic gap (item \texttt{mamo\_complex\_53}):} the vanilla MiMo-V2.5 solver assumed direct surplus-to-deficit transfers only, while the problem statement allowed indirect routes through intermediate regions. The DeepSeek-V4 critic caught the assumption explicitly (``the model restricts shipments to direct transfers from surplus to deficit regions only, but the problem allows arbitrary transfers including indirect routes''). The solver expanded the variable set on iteration~$2$ and produced $\phi = 13079$.
\end{enumerate}

The critic's interventions cluster into three failure modes: (i) execution failure (item~1), (ii) constraint completeness or type errors (items~2 and 3), and (iii) high-level formulation gaps (item~4). Cross-model pairings appear to be especially strong on (ii) and (iii), where the critic catches semantic mismatches that the solver's own distribution would judge as in-distribution. This is consistent with the rescue counts in \Cref{tab:rescue-counts} and with the asymmetry observed on A$^{2}$utoLPBench~s40x40, where failures are dominated by numerical precision (a regime where same-distribution agreement is more reliable than cross-distribution disagreement).

\subsection{Failure Analysis: OpenAI-GPT5.4 Regression Under A$^{2}$utoLPBench}
\label{app:gpt-regression}

The multi-solver leaderboard in \Cref{tab:leaderboard_xmodel} identifies one regression: OpenAI-GPT5.4 drops from $15.09\%$ vanilla to $13.21\%$ under A$^{2}$utoLPBench with the MiMo-V2.5 critic. To understand why, we audit the per-item iteration history.

\paragraph{Aggregate behaviour.}
On the 53-item MAMO complex subset, the MiMo-V2.5 critic disagreed at least once on $14$ items. Of those, only $2$ ($14\%$) were eventually returned correctly by OpenAI-GPT5.4 after the disagreement; the remaining $12$ either executed but produced the same numerical answer as the disagreed code, or produced new code that crashed with empty stdout. The iteration distribution skews toward $K{=}1$ ($39$ items, no disagreement) with a long tail of $K{=}2$ ($7$) and $K{=}3$ ($7$). OpenAI-GPT5.4 thus rarely engages the Refine path, and when it does not robustly incorporate the critic's feedback.

\paragraph{Item-level pattern.}
\Cref{tab:gpt-regression} summarises three representative failure modes; see the listed instance IDs for full traces in the released JSONL log.

\begin{table}[!htb]
    \centering
    \caption{Three representative behaviour patterns of OpenAI-GPT5.4 under MiMo-V2.5 critic feedback on MAMO complex.}
    \label{tab:gpt-regression}
    \small
    \setlength{\tabcolsep}{4pt}
    \begin{tabular}{p{0.3\linewidth} p{0.65\linewidth}}
        \toprule
        \textbf{Pattern} & \textbf{Trace} \\
        \midrule
        \texttt{mamo\_complex\_43}: vanilla correct, A$^{2}$LP wrong & Vanilla OpenAI-GPT5.4 returns $\phi = 5213$ on iteration~1 (matching the certified ground truth). Under A$^{2}$utoLPBench the prepended LP-optimization specification produces a different code that fails to execute (empty stdout); the critic, seeing no candidate output, returns \textsc{Agree} by default and the loop terminates without a recovery. Net: a true positive in vanilla becomes an execution failure in A$^{2}$utoLPBench. \\
        \midrule
        \texttt{mamo\_complex\_7}: persistent numerical near-miss & Critic disagrees on iteration~1 (``the candidate uses only Food\_4, providing only $6$~grams of protein per unit''), OpenAI-GPT5.4 rewrites; iteration~2 still emits $12.33$ vs.\ certified $13.0$ ($5\%$ relative error). Critic accepts the second answer because the formulation is now broadly plausible, but the numerical solution is still outside the $1\%$ tolerance. \\
        \midrule
        \texttt{mamo\_complex\_8}: feedback ignored & Critic disagrees with a specific structural diagnosis (``the candidate uses only Rice to meet all nutritional requirements; Rice alone provides only $X$~grams of protein''). OpenAI-GPT5.4's iteration~2 retains the same single-ingredient strategy with a slightly different magnitude ($9.7$ vs.\ certified $10.0$). The critic AGREEs because the answer is now numerically close, but the underlying formulation gap is unchanged. \\
        \bottomrule
    \end{tabular}
\end{table}

\paragraph{Interpretation.}
Two structural observations are visible across these traces. First, OpenAI-GPT5.4 with \texttt{reasoning\_effort=none} produces sound Propose-stage formulations efficiently but Refines them shallowly: the second iteration tends to perturb numerical coefficients without changing the constraint structure, even when the critic identifies a structural defect. Second, the prepended LP-optimization specification adds $\sim 1{,}200$ tokens to the system prompt, and on \texttt{mamo\_complex\_43} this shift was sufficient to push OpenAI-GPT5.4 onto a different (incorrect) Propose-stage code path even before the critic engaged.

The bundled solver-critic loop therefore exposes a capability threshold: the loop only adds value when the underlying solver can correctly act on a critic's structural diagnosis. When the solver's Refine policy is shallow (numerical perturbation rather than structural correction), critic feedback either fails to fix the error or, in rare cases, perturbs an already-correct vanilla Propose output. This finding is consistent with the rescue/regression asymmetry of \Cref{tab:rescue-counts}: stronger solvers (DeepSeek-V4, Claude-Sonnet-4.6, Kimi-K2.5) yield larger net rescue counts, while weaker or low-reasoning configurations yield near-zero net effect or, in the OpenAI-GPT5.4 case, a small negative.

\clearpage
\section{Extended Related Work}
\label{app:related-work}

\Cref{sec:intro} sketches A$^{2}$utoLPBench's relationship to four lines of prior work: LP-from-text benchmarks, contamination-resistant evaluation, generator-based reasoning benchmarks, and solver-critic / self-refinement protocols. We expand each line here.

\paragraph{LP-from-text and OR-LLM benchmarks.}
Early work in LP-from-text introduced human-curated word-problem collections paired with hand-labeled mathematical formulations or optimal values. NL4OPT~\cite{ramamonjison2022nl4opt} popularised the natural-language-to-LP formulation task on a $1{,}101$-item corpus; MAMO~\cite{huang2024mamo} and OptiBench~\cite{yang2024optibench} extended evaluation to end-to-end solver-code generation on similar-scale corpora, with OptiBench measuring an LLM's ability to call an external solver and return the numerical optimum. The OPTIMUS series~\cite{ahmaditeshnizi2023optimus,ahmaditeshnizi2024optimus,ahmaditeshnizi2024optimusv03} packages a multi-agent pipeline that develops a model, writes solver code, evaluates the candidate, and self-improves; OptiMUS-0.3 reports state-of-the-art accuracy on (mixed-integer) LP from natural-language descriptions. Recent reasoning-LLM agents, including OR-LLM-Agent~\cite{zhang2025or} and OptimAI~\cite{thind2025optimai}, push this line by combining stronger solver agents with explicit reasoning traces. All of these systems evaluate against static-corpus test sets in the sense of \Cref{eq:benchmark_def}: the released $\mathcal{D}$ is a fixed file with $\phi_i$ obtained from a solver run or a human annotator. A$^{2}$utoLPBench differs in two structural respects: it ships a generator rather than a corpus, and it certifies $\phi$ directly by the KKT theorem (\Cref{thm:optimality}) without any solver invocation or human annotation step.

\paragraph{Generator-based and contamination-resistant evaluation.}
DyVal~\cite{zhu2023dyval} samples directed acyclic graphs of varying sizes, converts each graph to a natural-language reasoning task, and evaluates an LLM by querying the value of a designated node; difficulty is controlled by the graph parameters. NPHardEval~\cite{fan2024nphardeval} runs the same recipe on classical NP-hard problems (TSP and others), parameterizing by problem size and complexity class. Both demonstrate that fresh seed ranges chosen after a model's training cutoff cannot have leaked into the training corpus, which is the structural guarantee A$^{2}$utoLPBench inherits. The broader contamination literature~\cite{mirzadeh2024gsm,zhang2024careful,xu2024benchmark,sainz2023nlp,carlini2021extracting,gao2020pile} documents inflated accuracy on memorised test items; recent surveys synthesise this evidence with detection and decontamination protocols~\cite{chen2025recent}. Among prior systems, A$^{2}$utoLPBench is closest in spirit to DyVal and NPHardEval (parameterized generation, contamination-by-construction), and closest in subject to NL4OPT/MAMO/OptiBench (LP-from-text); the two threads have not previously been combined.

\paragraph{Solver-critic and self-refinement protocols.}
The solver-critic loop of \Cref{sec:method-dualagent} is closely related to a line of self-refinement work. Self-Refine~\cite{madaan2023self} and Reflexion~\cite{shinn2023reflexion} have a single LLM iteratively generate, critique, and revise its own output, using verbal feedback in place of weight updates; CRITIC~\cite{gou2023critic} replaces the self-feedback with tool-grounded feedback (code interpreters, search engines), and multi-agent debate~\cite{du2024improving} replaces it with peer agents arguing different positions. Our protocol uses a separate critic agent on the same problem statement, which is most similar to multi-agent debate restricted to two participants. Importantly, recent critical surveys~\cite{huang2023large,kamoi2024can} document that internal self-correction often fails or even degrades performance when the underlying task is already saturated for the solver, with success requiring some form of grounded external signal. \Cref{tab:bigtable}'s same-distribution A$^{2}$LP MiMo-V2.5+MiMo-V2.5 underperforming vanilla on saturated NL4OPT (a $7.5$pp regression) is precisely an instance of that pattern; the empirical bound check in \Cref{app:dualagent} formalises both directions of critic noise (false negatives and false positives) as the mechanism behind the gap.

\paragraph{Verifiable reasoning and process supervision.}
A complementary line treats correctness as a property to be verified step-by-step. Math-Shepherd~\cite{wang2024math} trains a process reward model that scores each intermediate reasoning step on math problems, breaking the reliance on outcome-only reward; rStar-Math~\cite{guan2025rstar} pairs process supervision with Monte Carlo Tree Search to elicit deep reasoning in small models. Faithful chain-of-thought~\cite{lyu2023faithful}, Program-of-Thoughts~\cite{chen2022program}, and program-aided language models~\cite{gao2023pal} delegate the verifiable computation to a deterministic interpreter rather than the language model. A$^{2}$utoLPBench's KKT certificate is a structural counterpart of these mechanisms in the LP setting: the optimum is verified once, in closed form, at construction time rather than as a learned reward. Inverse optimization~\cite{chan2025inverse} provides the mathematical foundation for the construction direction we use; our inverse-KKT generator instantiates the duality / complementary-slackness route to recovering an LP from a chosen optimum (see \Cref{sec:method-generation}).

\paragraph{Agent-runnable benchmarks and tool-use.}
Recent agent benchmarks pair concrete environments with end-to-end task evaluation: AgentBench~\cite{liu2023agentbench} spans eight environments (OS, database, web, etc.); SWE-bench~\cite{jimenez2023swe} evaluates GitHub-issue resolution; ToolBench/ToolLLM~\cite{qin2023toolllm} measures large-scale API selection and invocation; WebArena~\cite{zhou2023webarena}, OSWorld~\cite{xie2024osworld}, and $\tau$-bench~\cite{yao2024tau} push toward realistic web, OS, and tool-agent-user interaction settings. These benchmarks inherit static-corpus structure on the task side: tasks are released once and become fixed targets. A$^{2}$utoLPBench's Docker runtime (\Cref{app:docker}) is most directly comparable to the agent-runnable surface these benchmarks ship; the difference is that the underlying tasks themselves are generator-produced, so a downstream evaluator can request fresh contamination-free batches at any time without coordinating with the benchmark authors.

\clearpage
\section{Agent-Friendly Docker Environment and Usage Guide}
\label{app:docker}

To make A$^{2}$utoLPBench convenient for downstream agents and researchers, we ship a self-contained Docker image (\texttt{autolpbench:latest}) built on Ubuntu 22.04.
The image bundles the inverse-KKT generator, scipy/PuLP solver libraries, the bundled solver-critic runtime, the 256-instance reference snapshot, and an \texttt{AGENTS.md} usage manual targeted at LLM-driven agents.
A downstream agent can attach to the image with one command:

\begin{lstlisting}[style=promptblock]
docker run --rm \
  -e LLM_PROVIDER=deepseek \
  -e LLM_API_KEY=$YOUR_KEY \
  autolpbench:latest \
  --mode autolpbench --strata s5x6,s10x12
\end{lstlisting}

The container exposes four entry points:
(i) generating fresh batches with arbitrary $(n,m,\text{seed-range})$ for contamination-free evaluation;
(ii) running vanilla or AutoLPBench-mode inference against any OpenAI- or Anthropic-compatible API;
(iii) re-verifying any released item with scipy;
(iv) returning per-item logs in JSONL alongside an aggregate summary.
The image is built with Python 3.11 and pinned versions of \texttt{numpy}, \texttt{scipy}, \texttt{pulp}, \texttt{openai}, and \texttt{anthropic}; total size is approximately $500$\,MB.

\paragraph{AGENTS.md and tool-use skill plug-and-play.}
The image ships its own usage manual at \texttt{/workspace/AGENTS.md}. The manual is written for LLM-driven agents to consume directly: it declares a fixed set of tool-use skills that the container exposes, each described in the JSON-schema format consumed by current tool-use APIs (OpenAI function calling, Anthropic tools, etc.). Any agent that supports tool use can read this manifest, register the skills, and run the benchmark without writing new glue code.

The four declared skills are:
\texttt{fetch\_problem(stratum, index)} which returns the natural-language description $T$ and the matrix-shape hints $(n, m)$;
\texttt{solve(code)} which executes a candidate Python program against the sandbox and returns its stdout and a parsed \texttt{FINAL\_ANSWER};
\texttt{verify(answer)} which compares a candidate answer to the certified $\phi$ and returns a Boolean plus the relative error;
and \texttt{generate\_batch(n, m, seed\_range)} which produces fresh contamination-free instances on demand.
A representative excerpt of the manifest is shown below; the full manifest is included in the released image and at the anonymous repository URL.

\begin{lstlisting}[style=agentsmd, caption={Excerpt from \texttt{AGENTS.md} (the full manifest declares all four skills).}, captionpos=b]
# AGENTS.md: A^2utoLPBench tool-use manifest

## Available skills
The Docker image exposes the following tool-use skills.
Any agent that supports tool use can register them directly.

- name: fetch_problem
  description: Return the next LP word problem in the chosen stratum.
  parameters:
    type: object
    properties:
      stratum: { type: string, enum: [s2x3, s3x4, ..., s40x40] }
      index:   { type: integer }
    required: [stratum, index]
  returns: { text: string, n: integer, m: integer }

- name: solve
  description: Execute candidate Python code in the sandbox.
  parameters:
    type: object
    properties:
      code: { type: string }
    required: [code]
  returns: { stdout: string, final_answer: number }

# ... verify, generate_batch follow the same schema.
\end{lstlisting}

This makes the deployment self-describing: an agent that has never seen A$^{2}$utoLPBench before can pull the image, read \texttt{AGENTS.md}, register the four skills, and complete the benchmark without external glue code or manual configuration.

\paragraph{Why the Docker integration matters for a generator-based benchmark.}
A static-corpus benchmark only needs to ship a JSONL file.
A generator-based benchmark must ship the generator together with a deterministic execution environment, otherwise downstream evaluators cannot reproduce the cross-batch calibration property.
By packaging everything into a single image with pinned dependencies, we make the freshness and reproducibility guarantees of \Cref{tab:properties} actually obtainable in practice rather than promised in principle.
The same image powers the cross-batch calibration experiment in \Cref{app:scaling_envelope}: anyone with the image can regenerate the figure by running the entry-point with three different seed ranges.

\end{document}